\documentclass[conference]{IEEEtran}
\IEEEoverridecommandlockouts
\usepackage{cite}
\usepackage{amsmath,amssymb,amsfonts}
\usepackage{algorithmic}
\usepackage{graphicx}
\usepackage{textcomp}
\usepackage{xcolor}

\usepackage{algorithm}
\usepackage{booktabs}
\usepackage{multirow}
\usepackage{makecell}
\usepackage{threeparttable}
\newtheorem{definition}{Definition}

\newcommand{\Pcal}{{\mathcal{P}}}

\newcommand{\R}[1]{\mathbb{R}^{#1}}
\newcommand{\E}[1]{\mathbb{E}[{#1}]}
\DeclareMathOperator{\subjectto}{s.t.}
\newcommand{\SPD}{{\tt SPD}}
\newcommand{\EOD}{{\tt EOD}}

\definecolor{red}{RGB}{255, 0, 0}

\definecolor{blue}{RGB}{0, 0, 255}

\def\BibTeX{{\rm B\kern-.05em{\sc i\kern-.025em b}\kern-.08em
    T\kern-.1667em\lower.7ex\hbox{E}\kern-.125emX}}
\begin{document}

\title{Robust Server Defense Against Unreliable Clients in One-Shot Fair Collaborative Machine Learning
%
}


\author{
	\IEEEauthorblockN{Chia-Yuan Wu\IEEEauthorrefmark{1}, Frank E. Curtis\IEEEauthorrefmark{2}, Daniel P. Robinson\IEEEauthorrefmark{3}
    }
\IEEEauthorblockA{\IEEEauthorrefmark{1}\textit{Department of Industrial and Systems Engineering}, 
    \textit{Lehigh University}, 
    Bethlehem, PA, USA \\
    Email: chw222@lehigh.edu}
\IEEEauthorblockA{\IEEEauthorrefmark{2}\textit{Department of Industrial and Systems Engineering}, 
    \textit{Lehigh University}, 
    Bethlehem, PA, USA \\
	Email: frank.e.curtis@lehigh.edu}
\IEEEauthorblockA{\IEEEauthorrefmark{3}\textit{Department of Industrial and Systems Engineering}, 
    \textit{Lehigh University}, 
    Bethlehem, PA, USA \\
	Email: daniel.p.robinson@lehigh.edu}
}

\maketitle

\begin{abstract}
Collaborative machine learning (CML) enables multiple clients to train a global model jointly in a data-distributed setting. To address data privacy and communication efficiency, one-shot CML has been increasingly adopted, where clients communicate with the server only once by sharing synthetic or processed proxy data. This single-round communication, however, eliminates the possibility of iterative correction at the server, making the learning process particularly vulnerable to client unreliability. In this setting, unreliable clients, whether malicious or non-malicious, may provide biased proxy data that favors certain groups, thereby degrading the fairness of the global model and harming minority or unprivileged groups. In this work, we propose a server-side defense framework based on a bilevel optimization formulation. The proposed approach learns client-level weights to mitigate the influence of biased client proxy data while enforcing fairness constraints by using a very small trusted root dataset available at the server. Experimental results on benchmark datasets show that our method improves fairness with little accuracy loss under biased proxy data contributions from unreliable clients. Moreover, the proposed approach remains effective even when unreliable clients make up a majority of the system, consistently outperforming other existing methods.
\end{abstract}


\section{Introduction}

Collaborative machine learning (CML) is a
distributed learning approach that enables multiple clients to collaborate toward a shared objective~\cite{laal2012collaborative}. Clients learn a global model jointly that achieves high accuracy across clients by leveraging information from other clients’ data. To prioritize data privacy in CML, clients may share local models with the server, as in federated learning (FL)~\cite{mcmahan2017communication}, or share synthetic or processed data~\cite{wu2026bilevel}, while keeping local datasets private. Fairness has become a critical challenge in CML~\cite{vucinich2023current,huang2024federated}. Due to the data heterogeneity of clients, the global model may perform well on some clients and poorly on others, which can result in the unfair treatment of certain groups. To address this issue, various fair CML methods have been proposed to promote fairness explicitly by incorporating constraints or implicitly by adapting the training process (e.g., by adjusting the aggregation weights based on local fairness measures)~\cite{ezzeldin2023fairfed, salazar2023fair, rasouli2020fedgan}. 

To achieve communication efficiency and privacy preservation in CML, one approach is to employ one-shot CML~\cite{holland2024one}, where clients communicate with the server only once by sharing either fully trained local models~\cite{guha2019one} or processed proxy datasets, such as distilled~\cite{holland2024one} or synthetic datasets~\cite{wu2026bilevel}, rather than exchanging model updates over multiple rounds. Such proxy datasets can be obtained through techniques including dataset distillation~\cite{wang2018dataset}, data transformation~\cite{calmon2017optimized}, or synthetic data generation~\cite{wu2026bilevel}. Although this single-round method substantially reduces communication costs, the server no longer has the opportunity to iteratively correct the global model for robustness through communication with the clients.

The success of CML fundamentally relies on the assumption that participating clients faithfully adhere to the training protocol. (If some clients behave unreliably, vulnerabilities are introduced that can harm the performance of the global model.) In practice, this assumption is often violated by \textit{unreliable clients} who compromise the global model's performance either intentionally or unintentionally~\cite{ma2021federated}. This unreliability encompasses a broad spectrum of behaviors. For example, non-malicious clients may fail to contribute high-quality processed data due to unstable connections or computational limitations~\cite{bouacida2021vulnerabilities}. In contrast, malicious clients actively disrupt the system; this category includes lazy participants who minimize their effort to save resources~\cite{ma2022federated}, selfish clients motivated by personal  gain~\cite{xu2026idfl}, and clients that launch explicit adversarial attacks. In standard CML, such attacks often involve model poisoning, where clients manipulate model gradients or parameters~\cite{fang2020local, zhou2021deep, zhang2022fldetector}, and data poisoning, where clients tamper with their local datasets~\cite{tolpegin2020data, doku2021mitigating}. Such behaviors can severely degrade the fairness of the global model and disproportionately harm minority or unprivileged groups. As a result, the server should defend against such intentional or unintentional client unreliability during the aggregation phase.


\begin{figure}[t]
  \centering
  \includegraphics[width=\columnwidth]{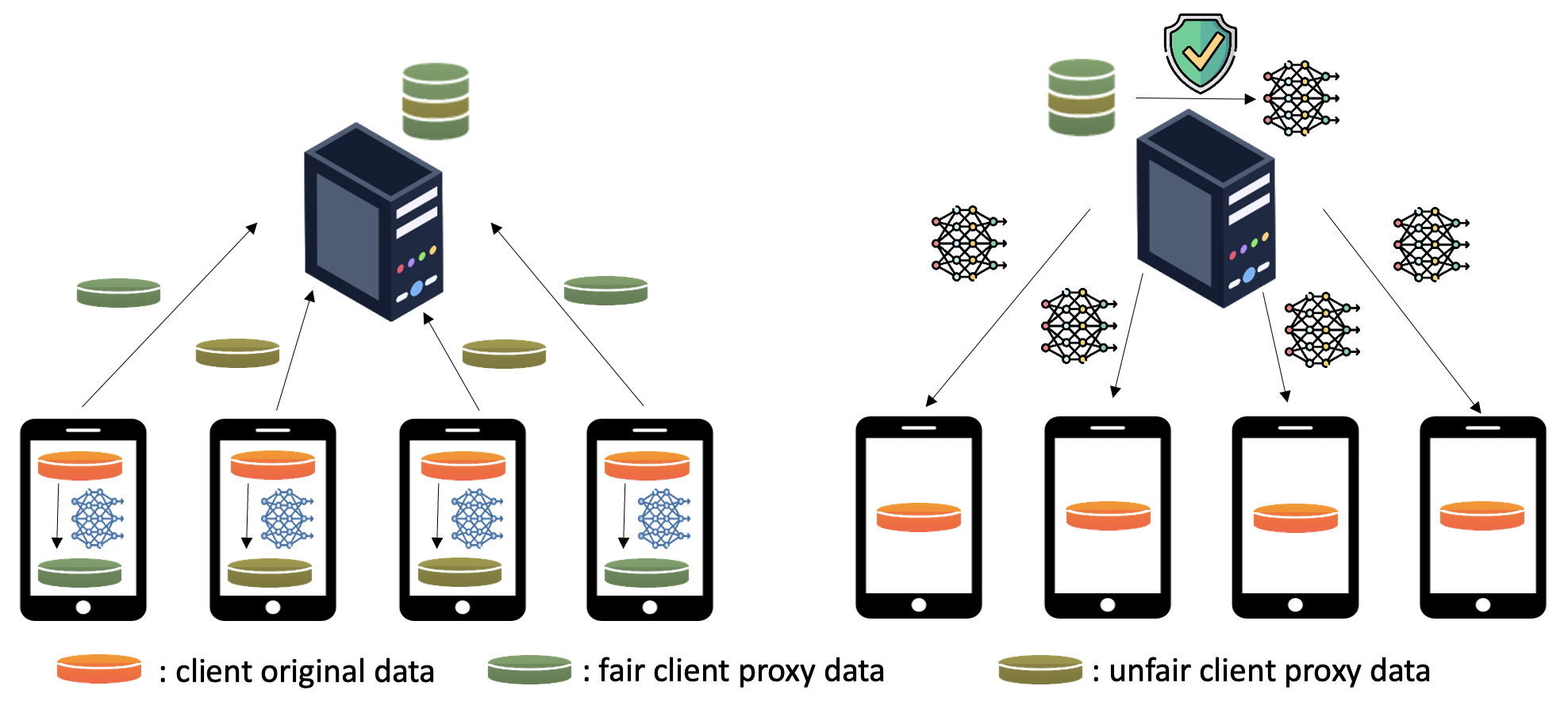}
  \caption{Illustration of the one-shot CML scenario considered in this paper, where the server receives some biased proxy data from unreliable clients and performs server-side defense. Icons are obtained from~\cite{flaticon}.}
  \label{fig:scenario}
\end{figure}

Regardless of whether originating from a malicious attack specifically targeting fairness or from non-malicious carelessness, the server receives biased proxy data and should perform server-side defense to mitigate the impact of such clients (see Figure~\ref{fig:scenario}). Motivated by this scenario, our work focuses on mitigating such biased client contributions.

Existing defense strategies in CML have been extensively studied in the standard multi-communication setting, and include robust aggregation rules~\cite{blanchard2017machine, yin2018byzantine}, weighting-based schemes~\cite{yi2022robust, cao2020fltrust, chen2023flram}, and anomaly detection and filter mechanisms ~\cite{chen2023flram, yi2022robust, jeong2022fedcc} that aim to reduce the influence of unreliable or adversarial clients. These approaches rely on iterative model updates, which allow the server to identify and mitigate problematic client contributions over multiple communication rounds. However, these assumptions do not hold in the one-shot CML setting considered in this work, where clients communicate with the server only once and no iterative correction is possible. As a result, biased data from unreliable clients can permanently affect the global model learned. Moreover, existing one-shot CML research primarily focuses on communication efficiency~\cite{guha2019one} or statistical heterogeneity~\cite{diao2023towards}. While some recent studies handle extreme statistical heterogeneity or incorporate feature alignment to prevent model performance degradation in the single-round setting \cite{zeng2024one, heinbaugh2023data}, they do not explicitly address fairness degradation. This gap highlights the need for a server-side defense mechanism designed for one-shot CML.

To bridge this gap, we present a server-side weighting-based approach for the one-shot CML setting to handle unreliable clients. We summarize our main contributions as follows:

\begin{itemize}
    \item We formulate a bilevel optimization problem that allows the server to learn client-level weights while incorporating fairness considerations into the optimization process.
    \item We validate our approach empirically on benchmark datasets from the literature. The results demonstrate its effectiveness in countering biased data contributions from unreliable clients in one-shot CML, overall achieving better fairness with minimal accuracy loss.
    \item Even when unreliable clients dominate the system (e.g., over 50\% of the clients), our method effectively mitigates unfairness while maintaining high accuracy.
\end{itemize}

\section{Background and Related Work}

In this section, we discuss the relevant background and related work on one-shot CML, fairness in CML, robustness and defense in CML, and bilevel optimization.

\subsection{One-Shot CML}

One-shot CML is a collaborative learning paradigm in which clients communicate with the server only once during the training process, thus reducing communication cost and system overhead compared to a multi-round framework~\cite{holland2024one}. In this setup, clients independently perform local training and transmit learned information in a single round, and then the server constructs a global model without further client participation. This single-round communication may involve fully trained local models, compressed representations, or synthetic proxy data depending on the system design and privacy requirements~\cite{amato2025towards}. Since one-shot CML lacks a multi-round communication process, it loses the opportunity for the server to (iteratively) refine or correct biased or low-quality client information. Consequently, the quality of the aggregation process is sensitive to the heterogeneity, bias, and quality of the client data.
Thus, many one-shot CML methods place greater emphasis on server-side aggregation strategies.

To address these challenges, existing work has proposed aggregation mechanisms tailored to one-shot CML/FL. For instance, early work on one-shot FL leverages ensemble learning and distillation to aggregate fully trained client models in a single communication round, aiming to improve performance while maintaining communication efficiency~\cite{guha2019one}. In~\cite{heinbaugh2023data}, a data-free generative approach is used to mitigate extreme statistical heterogeneity, allowing clients to share class-conditional knowledge without relying on auxiliary public datasets. In~\cite{zeng2024one}, heterogeneous client knowledge is preserved by treating local models as independent experts and combining them through an aggregation scheme. Despite these advances, fairness degradation caused by unreliable client contributions remains largely unexplored in the context of one-shot CML.

\subsection{Fairness in CML}\label{sec:fairness}

Fairness in CML can be broadly categorized as either \textit{group fairness} or \textit{client fairness}~\cite{chen2023privacy}. Group fairness focuses on mitigating bias in the global model’s predictions across different demographic groups, ensuring that the model learned does not disadvantage groups. In contrast, client fairness aims to prevent the global model from favoring or being overly influenced by certain clients. In this work, we address both group fairness and client fairness, with the objective of ensuring that the global model does not discriminate against a group defined by attributes such as race, gender, or age~\cite{mehrabi2021survey}, and is not negatively affected by unreliable clients. To quantify group fairness, existing studies commonly adopt \textit{statistical parity} (SP), which requires model predictions to be independent of sensitive attributes, and \textit{equal opportunity} (EO), which enforces equal true positive rates (TPR) across different groups, as evaluation criteria~\cite{pessach2022review}. In CML, violations of group fairness often arise because the server passively inherits biases embedded in the shared information from clients, regardless of whether these biases originate from unfair local data or are intentionally introduced during client-side information processing~\cite{benmalek2024bias}. Therefore, addressing fairness in a distributed learning setting requires the server to identify and correct biased client contributions during aggregation.

Ensuring fairness in CML is particularly challenging due to data heterogeneity across clients and the decentralized nature of local training and data ownership. To address these challenges, existing work utilizes iterative refinement to mitigate unfairness and progressively correct bias throughout the training process. Some approaches focus on fairness-aware aggregation by accounting for local group fairness performance (e.g.,  FairFed~\cite{ezzeldin2023fairfed} and FAIR-FATE~\cite{salazar2023fair}), while others explore post-processing-based strategies where clients locally apply fairness corrections after the global model has converged, thereby decoupling fairness enforcement from the distributed training process~\cite{zhou2025post}. In addition, optimization-based frameworks such as GLOCALFAIR incorporate group fairness constraints at both the client and server levels through iterative client-server coordination~\cite{meerza2024glocalfair}. Despite these efforts, most fairness-aware CML methods rely on multi-round communication and iterative correction, which makes them fundamentally incompatible with \emph{one-shot} CML.

\subsection{Robustness and Defense in CML}

CML relies on clients following the prescribed training protocol and contributing reliable information to support the global objective. In practice, this assumption is often violated because of unreliable clients that contribute corrupted, low-quality, or strategically manipulated data~\cite{ma2021federated}. Such behavior, whether malicious or not, can substantially degrade the utility and robustness of the learned global model if the server blindly aggregates all client contributions. To address this issue, prior work has studied server-side defense mechanisms in CML/FL, primarily in the multi-round communication setting. Existing approaches can be broadly categorized into three classes: \textit{robust aggregation}, \textit{anomaly detection and client filtering}, and \textit{trust- or weighting-based} methods.

Robust aggregation methods enhance robustness by selecting local models that exhibit similar behaviors, such as comparable loss values, proximity to the median, or consistency with the majority of clients~\cite{blanchard2017machine, yin2018byzantine, talukder2022computationally, le2025fednolowe}. By filtering out clients whose contributions deviate significantly from these patterns, these methods mitigate the impact of outliers and adversarial updates.
Anomaly detection and client filtering techniques explicitly identify and remove suspicious client updates before aggregation. For example, FedCC~\cite{jeong2022fedcc} leverages feature-level similarity to cluster benign and malicious clients, while FLRAM~\cite{chen2023flram} combines anomaly detection, clustering, and credibility scoring to filter abnormal updates. Both are effective against various poisoning and backdoor attacks by detecting inconsistent client behavior. Trust- or weighting-based methods assign aggregation weights according to estimated client reliability or contribution quality. Prior work also explores sample-level reweighting using clean validation sets or meta-learning to mitigate noise and imbalance~\cite{ren2018learning}, as well as client-level adaptive weighting schemes that leverage gradient similarity, loss variance, or pairwise model similarity to stabilize training under data heterogeneity~\cite{wu2021fast, reyes2021precision, liu2023feddwa, talukder2022computationally, le2025fednolowe}. Recently, methods such as FedLAW learn client aggregation weights with global regularization to promote coherent updates and improve generalization in non-IID settings~\cite{li2023revisiting}. Furthermore, trust-based weighting methods such as FLTrust incorporate a small trusted root dataset to evaluate client updates and suppress unreliable contributions~\cite{cao2020fltrust, yi2022robust, geng2023better}. 

Despite their effectiveness, existing robustness-based reweighting methods implicitly assume that unreliable clients remain in the minority, allowing majority-based statistics (e.g., median) to effectively identify and suppress outliers. When unreliable clients dominate the system, these assumptions break down and such methods may fail to provide adequate protection. Moreover, these approaches depend on iterative observations and assume the ability to intervene across multiple communication rounds. However, these assumptions no longer hold in the one-shot CML setting, which motivates the need for alternative server-side mechanisms that can handle unreliable clients in the single-round communication case.

\subsection{Bilevel Optimization}

Bilevel optimization involves hierarchical optimization problems in which one (the inner problem) is embedded within another (the outer problem) through a constraint. The standard bilevel optimization formulation can be expressed as
\begin{equation}
\begin{aligned}
\min_{x \in \R{n},\ y \in \R{m}} \
 &F_o(x,y) \\
 \ \text{subject to (s.t.)} \ \
&x \in X,\ y \in \arg\min_{y \in Y(x)} F_i(x, y),
\end{aligned}
\end{equation}
where $F_o$ and $F_i$ denote the objective functions of the outer (upper level) problem and the inner (lower level) problem~\cite{savard1994steepest}. In this hierarchical structure, the upper-level decision variable $x$ influences the feasible set $Y(x)$ as well as the optimal solution and objective value of the lower-level problem~\cite{sinha2017review}. Due to this nested dependency, the optimal solution of the inner problem, denoted as $y$, affects the upper-level optimal solution and objective value~\cite{giovannelli2024bilevel}.

In this work, we adopt a bilevel optimization framework as a modeling tool to support the proposed server-side defense mechanism. The detailed formulation and algorithmic design of our approach are presented in the next section.


\section{Proposed Solution}

In this section, we describe a client-side proxy data generation process and propose a server-side defense framework.

\subsection{Client-side Proxy Data Generation}

In the one-shot CML setting considered in this paper, each client communicates with the server only once by transmitting a proxy dataset. This proxy dataset is intended to summarize the client’s local information while reducing communication cost and preserving data privacy. Since no further interaction with the server is allowed, all local data processing must be completed prior to this single communication round.

We do not restrict how clients generate their proxy datasets. In general, each client may adopt any fairness-aware data processing strategy to mitigate bias during proxy data generation, including data distillation~\cite{wang2018dataset}, data transformation~\cite{calmon2017optimized}, or synthetic data generation~\cite{wu2026bilevel}. Such proxy data is typically intended to correct unfairness in the original local dataset. However, due to unreliable client behavior, the fairness of the transmitted proxy data cannot be guaranteed. As a result, biased client contributions may still be passed to the server and directly affect the fairness of the learned global model.

Formally, each client $c \in K$ uses its original local dataset $\{d_{i,c}^o = (x_{i,c}^o, s_{i,c}^o, y_{i,c}^o)\}_{i=1}^{N_c^o}$ to generate a proxy dataset $\{d_{i,c}^p = (x_{i,c}^p, s_{i,c}^p, y_{i,c}^p)\}_{i=1}^{N_c^p}$, which is sent to the server. Here, $\{x_{i,c}^o,x_{i,c}^p\} \subset \R{n-1}$ are the non-sensitive features for the original and proxy datasets, $\{s_{i,c}^o,s_{i,c}^p\} \subset \{0,1\}$ are the sensitive features for the original and proxy datasets, $\{y_{i,c}^o,y_{i,c}^p\} \subset \{-1,1\}$ are the labels for the original and proxy datasets, $\{d_{i,c}^o,d_{i,c}^p\} \subset \R{n+1}$ are the complete data representation of the $i$-th sample from client $c$ for the original and proxy data, and $N_c^o$ and $N_c^p$ are the numbers of samples in the original and proxy datasets of client $c$, respectively. Algorithm~\ref{alg:client} summarizes this generic client-side proxy data generation procedure. The algorithm does not specify a particular instantiation of the proxy data generation method.  More specificity on what the server requires of the proxy data is given at the end of Section~\ref{subsec:server} (see Definition~\ref{def:fair-proxy-dataset}).

\begin{algorithm}[t]
\caption{Each Client $c$ Generates its Proxy Dataset}
\label{alg:client}
\small
\begin{algorithmic}[1]

\STATE \textbf{Input:} Client original dataset $\{d_{i,c}^o\}_{i=1}^{N_c^o}$ 
\STATE \textbf{Output:} Client proxy dataset
$\{d_{i,c}^p\}_{i=1}^{N_c^p}$ 

\STATE Choose desired proxy data size $N_c^p$
\STATE Generate a proxy dataset $\{d_{i,c}^p = (x_{i,c}^p,s_{i,c}^p,y_{i,c}^p)\}_{i=1}^{N_c^p}$ (e.g., using a fairness-aware method such as those proposed in~\cite{wu2026bilevel, calmon2017optimized}) 
\STATE Send $\{d_{i,c}^p\}_{i=1}^{N_c^p}$ to the server

\end{algorithmic}
\end{algorithm}


\begin{algorithm}[t]
\caption{Server-side Defense in One-shot CML}
\label{alg:server.2.p}
\small
\begin{algorithmic}[1]

\STATE \textbf{Input:} Proxy datasets from clients
$\{d_{i,c}^p\}_{i=1,c=1}^{N_c^p,K}$ obtained from Algorithm~\ref{alg:client}, and root datasets
$\{d_{i,c}^r\}_{i=1,c=1}^{N_c^r,K}$ for the clients

\STATE \textbf{Output:} Global model parameter vector  $\theta^G$

\FOR{each $t = 1,\dots,t_{\max}$}
    \STATE Compute $\nabla \mathcal{P}(w_t)$ of
    (\ref{prob:main.orig.server.2.p.obj})
    by using implicit differentiation
    \STATE Update $\tilde{w}_{t+1}$ using $w_t$ and
    $\nabla \mathcal{P}(w_t)$
    (e.g., using gradient descent or Adam) \label{line:gradient.P}
    \STATE Compute $w_{t+1}$ by projecting $\tilde{w}_{t+1}$ onto the standard simplex defined by   (\ref{prob:main.orig.server.2.p.c1}) and (\ref{prob:main.orig.server.2.p.c2})
    using Algorithm~\ref{alg:proj.simplex}
\ENDFOR

\STATE Compute 
$\theta^G$ as the solution to~(\ref{prob:global}) with weights $w_{t_{\max}+1}$ 

\STATE Broadcast global model parameter vector $\theta^G$ to all clients

\end{algorithmic}
\end{algorithm}


\begin{algorithm}[t]
\caption{Projection onto the Standard Simplex~\cite{duchi2008efficient}}
\label{alg:proj.simplex}
\small
\begin{algorithmic}[1]

\STATE \textbf{Input:} Tentative vector $\tilde{w}_{t} \in \mathbb{R}^{K}$

\STATE \textbf{Output:} Projected vector $w_{t}$ satisfying
(\ref{prob:main.orig.server.2.p.c1}) and
(\ref{prob:main.orig.server.2.p.c2})

\STATE Sort $\tilde{w}_{t}$ into $u$ such that $u_1 \ge u_2 \ge \dots \ge u_K$

\FOR{each $k = 1,\dots,K$}
    \STATE Compute
    $\tau_k = \frac{1}{k}
    \left(\sum_{i=1}^{k} u_i - 1\right)$
\ENDFOR

\STATE $\rho = \max \{ k \in K : u_k - \tau_k > 0 \}$
\STATE Set shifting value $\tau = \tau_{\rho}$
\STATE Compute $w_t = \max \{ \tilde{w}_{t} - \tau, 0 \}$
\STATE \textbf{return} $w_t$

\end{algorithmic}
\end{algorithm}

\subsection{Server-side Defense}\label{subsec:server}

Given the proxy datasets $\{d_{i,c}^p\}_{i=1,c=1}^{N_c^p,K}$ received from all clients, the server aims to learn a global model that is accurate and fair. However, due to unreliable client behavior, the proxy data transmitted to the server may still exhibit bias. In one-shot CML, the server has no opportunity to iteratively refine or correct client contributions through repeated communication rounds. As a result, directly aggregating all client proxy datasets can lead to unfairness in the learned global model.

To address this challenge, we propose a server-side defense framework based on bilevel optimization. The core idea is to learn client-level aggregation weights at the server, such that biased or unreliable client contributions are adaptively downweighted during aggregation. This allows the server to mitigate the influence of unreliable clients while promoting group fairness in the learned global model. Our proposal assumes that the server holds a small trusted root dataset $\{d_{i,c}^r = (x_{i,c}^r, s_{i,c}^r, y_{i,c}^r)\}_{i=1}^{N_c^r}$ for each client $c \in K$. This root dataset is assumed to be drawn from the same distribution as the client’s original local dataset and is kept very small (e.g., $0.5\%$ of $N_c^o$), serving as a lightweight held-out audit or validation dataset on the server. This reliance on a root dataset is a limitation that may not be feasible in all scenarios.

To incorporate group fairness into the optimization problem, we consider statistical parity (SP) and equal opportunity (EO) as defined in Section~\ref{sec:fairness}. Both metrics are expressed in terms of conditional probabilities of model predictions and are generally non-convex with respect to the model parameters $\theta$, posing significant challenges for direct optimization~\cite{caton2024fairness}. Following~\cite{zafar2019fairness}, we adopt the decision boundary covariance (DBC) as a tractable fairness approximation. In particular, for SP, the DBC is defined as \par\nointerlineskip
{\footnotesize
\begin{equation} \label{equ:DBC}
\begin{aligned}
\text{Cov}(S, d_{\theta}(X,S)) 
&= \E{(S-\E{S}) d_{\theta}(X,S)} - \E{S-\E{S}}\E{d_\theta(X,S)} \\
&= \E{(S-\E{S}) d_{\theta}(X,S)}, 
\end{aligned}
\end{equation}}\noindent
where $X$ denotes the random vector of non-sensitive features, $S$ the sensitive attribute, and $d_\theta(X,S)$ the signed distance to the decision boundary defined by $\theta$. The covariance measures the linear dependence between the sensitive attribute and the decision function. Enforcing this quantity to be close to zero diminishes the statistical dependency between model predictions and the sensitive attribute, thereby promoting fairness. A similar covariance-based surrogate is used for EO by conditioning the expectation on the positive label.

We then incorporate group fairness into our bilevel optimization framework by enforcing constraints on this covariance. Specifically, the server learns a global model \par\nointerlineskip
{\footnotesize
\begin{equation} \label{prob:global}
\theta^G := \theta_w\left(\{d_{i,c}^p\}_{i=1,c=1}^{N^p,K}, \{w_c^*\}_{c=1}^K\right),
\end{equation}}\noindent
where $\{w_c^*\}_{c=1}^K$ is obtained by solving the following bilevel optimization problem, where for notational simplicity we let $\{w_c^*\}:= \{w_c^*\}_{c=1}^K$, $\{w_c\}:= \{w_c\}_{c=1}^K$, $\{d_{i,c}^p\}:= \{d_{i,c}^p\}_{i=1,c=1}^{N_c^p, K}$, $\{d_{i,c}^r\}:= \{d_{i,c}^r\}_{i=1,c=1}^{N_c^r, K}$, and $a_{i,c}^r:= (x_{i,c}^r, s_{i,c}^r)$: \par\nointerlineskip
{\footnotesize
\begin{subequations}\label{prob:main.orig.server}
\begin{align}
\!\{w_c^*\}
&:= \arg\min_{\{w_c\}}
\ \ell_{N^r,w,r}\Big(\{d_{i,c}^r\}, \theta_w\big(\{d_{i,c}^p\},\{w_c\}\big)\Big)
\label{prob:main.orig.server.obj}
\\
\subjectto
& \sum_{c=1}^K w_c = 1 \label{prob:main.orig.server.c1}
\\
& w_c \ge 0, \quad \text{for all $c = 1,\dots,K$} \label{prob:main.orig.server.c2}
\\
& \left|
\frac{1}{N^R}
\sum_{c=1}^K \sum_{i=1}^{N_c^r} w_c(s_{i,c}^r - s_{avg}^r) (a_{i,c}^r)^{\!T}\theta_w\big(\{d_{i,c}^p\},\{w_c\}\big)\right|\le \epsilon^{SP}
\label{prob:main.orig.server.c.SP}
\end{align}
\end{subequations}}\noindent
with
{\footnotesize
\begin{align*}
N^R := \sum_{c=1}^K N_c^r, \quad s_{avg}^r := \frac{1}{N^R} \sum_{c=1}^K \sum_{i=1}^{N_c^r} s_{i,c}^r,
\end{align*}}
\noindent
\!\!function $\theta_w: \prod_{i=1}^{N} (\R{n-1}\times\{0,1\}\times\{-1,1\}\times\R{}) \to \R{n}$ is \par\nointerlineskip
{\footnotesize
\begin{align}\label{def:theta.w.function}
\theta_w\big(\{d_{i,c}\}, \{w_c\} \big)
:=
\arg\min_{\theta} \ \ell_{N,w,r} \Big(\{d_{i,c}\}, \{w_c\}, \theta \Big),
\end{align}} \noindent
\!\!and the loss function is \par\nointerlineskip
{\footnotesize
\begin{align}
&\ell_{N,w}(\{d_{i,c}\},\{w_c\},\theta) := \frac{1}{N} \sum_{c=1}^K w_c \sum_{i=1}^{N_c}  \log\left(1+e^{-y_{i,c} (a_{i,c})^T\theta}\right), \\
&\ell_{N,w,r}(\{d_{i,c}\},\{w_c\},\theta):= \ell_{N,w}(\{d_{i,c}\},\{w_c\},\theta) + reg(\theta),
\end{align}} \par \noindent
where $N := \sum_{c=1}^K N_c$ and $reg(\theta) := \frac{\lambda_\theta}{2n^2} \|\theta\|_2^2$. Alternatively, when EO is considered, we impose the following constraint, which serves as the EO counterpart to the SP constraint in~\eqref{prob:main.orig.server.c.SP}: \par\nointerlineskip
{\footnotesize
\begin{align}\label{prob:main.orig.server.c.EO}
\left|
\frac{1}{N^R} \sum_{c=1}^K \sum_{i=1}^{N_c^r} w_c (s_{i,c}^r - s_{avg}^r) m_{i,c}^r (a_{i,c}^r)^{\!T} \theta_w\big(\{d_{i,c}^p\},
\{w_c\} \big) \right| \le \epsilon^{EO},
\end{align}} \par \noindent
where $m_{i,c}^r := (1+y_{i,c}^r)/2$. Both constraints are empirical realizations of the DBC surrogate defined in equation~\eqref{equ:DBC}.

The bilevel optimization structure in problem~\eqref{prob:main.orig.server} captures the hierarchical interaction between client data aggregation and fairness enforcement in the proposed server-side defense mechanism. The inner problem applies the learned client-level weights $\{w_c\}$ to aggregate the proxy datasets $\{d_{i,c}^p\}$ and trains a classification model $\theta_w\big(\{d_{i,c}^p\},\{w_c\}\big)$ by minimizing the weighted regularized loss as defined in problem~\eqref{def:theta.w.function}. The outer problem then evaluates this resulting model on the trusted root dataset $\{d_{i,c}^r\}$ to optimize the client weights $\{w_c\}$ by minimizing prediction loss on the weighted root data, subject to the fairness constraints~\eqref{prob:main.orig.server.c.SP} (and/or~\eqref{prob:main.orig.server.c.EO}) and the simplex constraints~\eqref{prob:main.orig.server.c1} and~\eqref{prob:main.orig.server.c2}. This nested formulation allows the server to adaptively downweight unreliable clients whose proxy data undermines fairness, while maintaining model training on the reweighted aggregated data.

We adopted a penalty reformulation of problem~\eqref{prob:main.orig.server} that allows efficient gradient-based optimization while preserving the fairness regularization effect. The reformulated problem, where for notational simplicity we use $w := \{w_c\}_{c=1}^K$, is
\par\nointerlineskip
{\footnotesize
\begin{subequations}\label{prob:main.orig.server.2.p}
\begin{align}
w^* := \arg\min_{w} \ \
&\{\Pcal(w) = \ell_{N^r,r, w}(\{d_{i,c}^r\}, \theta_w(w)) \notag\\
&+ (1-\nu)\Pcal_{SP}(w) + \nu\Pcal_{EO}(w)\}  \label{prob:main.orig.server.2.p.obj}
\\
\ \subjectto \ 
 &\sum_{c=1}^K w_c = 1 \label{prob:main.orig.server.2.p.c1} \\
 & w_c \geq 0, \quad c=1,\dots,K, \label{prob:main.orig.server.2.p.c2}
\end{align}
\end{subequations}} \noindent
where
{\footnotesize
\begin{equation}
\begin{aligned}
\Pcal_{SP}(w)
&= \frac{\rho_s}{2}\left(\frac{1}{N^R} \sum_{c=1}^K  \sum_{i=1}^{N_c^r} w_c (s_{i,c}^r-s_{avg}^r) (a_{i,c}^r)^T \theta_w(w) \right)^2,
\\
\Pcal_{EO}(w)
&= \frac{\rho_s}{2}\left(\frac{1}{N^R} \sum_{c=1}^K  \sum_{i=1}^{N_c^r} w_c (s_{i,c}^r-s_{avg}^r) m_{i,c}^r (a_{i,c}^r)^T \theta_w(w) \right)^2,
\end{aligned}
\end{equation}
} \par \noindent
$\rho_s$ is the penalty parameter, $\nu \in [0,1]$ balances the relative importance between SP and EO, and the inner problem is \par\nointerlineskip
{\footnotesize
\begin{align}\label{prob-training}
\theta_w(w) := \arg\min_{\theta} \ell_{N^p,w,r}(\{d_{i,c}^p\}, w, \theta).
\end{align}}

The complete procedure to solve the proposed penalty-based reformulation~\eqref{prob:main.orig.server.2.p} is summarized in Algorithm~\ref{alg:server.2.p}. To optimize the outer objective $\Pcal(w)$, we compute its gradient with respect to $w$ via implicit differentiation of the inner problem. Specifically, the gradient is given by \par\nointerlineskip
{\footnotesize
\begin{equation}
\nabla \Pcal(w)
=
\left(
\frac{\partial \theta_w(w)}{\partial w}
\right)^{\!T}
\frac{\partial \Pcal(w)}{\partial \theta_w(w)} + \frac{\partial \Pcal(w)}{\partial w},
\end{equation}}\noindent
where $\frac{\partial \theta_w(w)}{\partial w}$ is obtained by differentiating the optimality conditions of the inner problem. This leads to the following linear system involving the Hessian of the inner objective:
 \par\nointerlineskip
{\footnotesize
\begin{equation}\label{eqa.dtheta.dw}
\left(
\frac{\partial^2 \ell_{N^p,w,r}(\{d_{i,c}^p\},w,\theta)}{\partial^2\theta}
\right)
\frac{d\theta_w}{dw}
= -  \frac{\partial^2 \ell_{N^p,w,r}(\{d_{i,c}^p\},w,\theta)}{\partial\theta\partial w}. 
\end{equation}}
\par \noindent
After each gradient update in line~\ref{line:gradient.P} in Algorithm~\ref{alg:server.2.p}, the tentative weight vector may violate the simplex constraints~\eqref{prob:main.orig.server.2.p.c1} and \eqref{prob:main.orig.server.2.p.c2}, so we project the updated weights onto the standard simplex. The projection is performed using the efficient sorting-based algorithm in~\cite{duchi2008efficient}, as detailed in Algorithm~\ref{alg:proj.simplex}.

Motivated by the bilevel optimization modeling formulation~\eqref{prob:main.orig.server}, the server requests that the proxy data it receives from each client is a \emph{fair} proxy dataset, as we now define. 

\smallskip
\begin{definition}[Fair Proxy Dataset]\label{def:fair-proxy-dataset}
A proxy dataset $\{d_{i,c}^p\}_{i=1}^{N_c^p}$ from client $c$ is called a \emph{fair proxy dataset} if the model parameters $\theta^C := \arg\min_\theta (1/N_c^p)\sum_{i=1}^{N_c^p}  \log\left(1+e^{-y_{i,c}^p (a_{i,c}^p)^T\theta}\right)$
satisfy the desired fairness criteria with respect to the client's original dataset (e.g., in the case of SP, $\theta^C$ satisfies the inequality $\left|
\frac{1}{N_c^o}
\sum_{i=1}^{N_c^o} (s_{i,c}^o - s_{avg}^o) (a_{i,c}^o)^{\!T}\theta^C\right|\le \epsilon^{SP}$.
\end{definition}


\smallskip
To be clear, the server \emph{requests} that every client supply a fair proxy dataset, but an unreliable client may fail to provide such a dataset. Our server-side defense strategy described in this section aims to handle such unreliable clients.

\section{Experiments}

\subsection{Datasets}

We consider two binary classification datasets commonly used in the fairness literature~\cite{salazar2023fair}: \texttt{Law School}~\cite{wightman1998lsac} and \texttt{Dutch}~\cite{van20012001}. The \texttt{Law School} dataset consists of 20,798 samples with 11 features, where the task is to predict whether a student will pass the bar exam, with \textit{race} as the sensitive attribute. The \texttt{Dutch} dataset contains 60,420 samples with 11 features, where the goal is to predict whether an individual holds a prestigious occupation, with \textit{sex} as the sensitive attribute. For both datasets, we partition the data among $K=5$ clients evenly and randomly, and each client further splits its local data into 80\% for training and 20\% for testing.

\subsection{Evaluation metrics}

We adopt three evaluation metrics. \textit{Accuracy} (Acc(\%)) measures the percentage of correctly classified samples. For evaluating fairness, we use the \textit{statistical parity difference} (SPD) and the \textit{equal opportunity difference} (EOD): \par\nointerlineskip
{\footnotesize
\begin{align*}
&\SPD{} \\
&= \frac{|\{i: y_\theta(x_i,s_i) = 1 \cap s_i=1\}|}{|\{i:s_i=1\}|} - \frac{|\{i: y_\theta(x_i,s_i) = 1 \cap s_i = 0\}|}{|\{i:s_i=0\}|}, \\ 
&\EOD{} \\
&= \frac{|\{i: y_\theta(x_i,s_i) = 1 \cap s_i=1 \cap y_i = 1\}|}{|\{i:s_i=1 \cap y_i = 1\}|} \\
& - \frac{|\{i: y_\theta(x_i,s_i) = 1 \cap s_i = 0 \cap y_i = 1\}|}{|\{i:s_i=0 \cap y_i = 1\}|},
\end{align*}} \par \noindent
where $y_\theta(x_i,s_i)$ denotes the predicted label for sample $i$. Values of $|\text{SPD}|$ and $|\text{EOD}|$ closer to zero indicate better fairness. Unlike typical distributed learning that evaluates overall system performance, we evaluate the global model only on the test data of reliable clients. This reflects our system objective: clients that adhere to the prescribed protocol should benefit from collaborative learning. Unreliable clients deviate from the protocol and transmit biased proxy data, so including their test data would obscure whether the defense mechanism effectively protects reliable participants.

\subsection{Implementation Details}\label{sec:implm.details}

The inner problem~\eqref{prob-training} is solved using the L-BFGS solver~\cite{wright2006numerical} with a convergence tolerance of $10^{-7}$ and a maximum of 1000 iterations. If the iteration limit is reached, the last iterate is used. For the outer problem, we use the Adam solver~\cite{kingma2014adam} with $t_{\max}=2000$ iterations 
and an initial learning rate of 0.1. The regularization parameter is set to $\lambda_\theta = 10^{-4}$, and the penalty parameter is adaptively adjusted. Specifically, we define a scheduled $\rho_s$ starting with a value of $10$ and increasing it by a factor of $10$ every $400$ iterations until it reaches $10^{4}$. 
We set $\nu \in \{0, 1\}$ in the penalty formulation~\eqref{prob:main.orig.server.2.p} to enforce SP-only or EO-only type constraints. The client aggregation weights are initialized uniformly.

\subsection{Experimental Scenarios}\label{subsec:scenarios}

We consider the unreliable client percentages 20\%, 40\%, and 60\%. The 60\% setting is especially challenging since the unreliable clients dominate the system. For each percentage, we design two experimental scenarios that we call \textit{ideal scenarios} and \textit{realistic scenarios}, which we describe next.

For the realistic scenarios, 
we first use each client's original training data to train the logistic loss function and compute the relevant fairness measure (i.e., $|\text{SPD}|$ or $|\text{EOD}|$) with respect to its original testing data, with the highest measure giving the most unreliable (unfair) client, the second highest being the second most unreliable client, etc. Then, for the $40\%$ unreliable client scenario as an example, we identify the $2$ most unreliable (unfair) clients ($2/5 = 0.4$) and send their original data to the server as their proxy data, and for the remaining $3$ clients (i.e., the $3$ most reliable (fair) clients), we apply the fairness-aware method proposed in~\cite{wu2026bilevel} to produce a fair proxy dataset of the same size as the original local dataset, i.e., $N_c^p = N_c^o$. A similar procedure is used for the $20\%$ and $60\%$ unreliable client scenarios. The above procedure is repeated $5$ times with different random seeds to obtain $5$ different data splits and corresponding scenarios, and the results we report are averaged across these $5$ runs.


For the ideal scenarios, let us consider for illustrative purposes the $40\%$ setting.  In this case, the $2$ unreliable clients and their proxy data are determined exactly as in the realistic scenario. The proxy datasets for the $3$ remaining reliable client datasets are chosen as replicas of each other and equal to the fair dataset computed using, again, the fairness-aware method proposed in~\cite{wu2026bilevel} applied to the dataset of the client determined to be the most reliable (fair), as described above.



\begin{figure}[t]
  \centering
  \includegraphics[width=0.9\columnwidth]{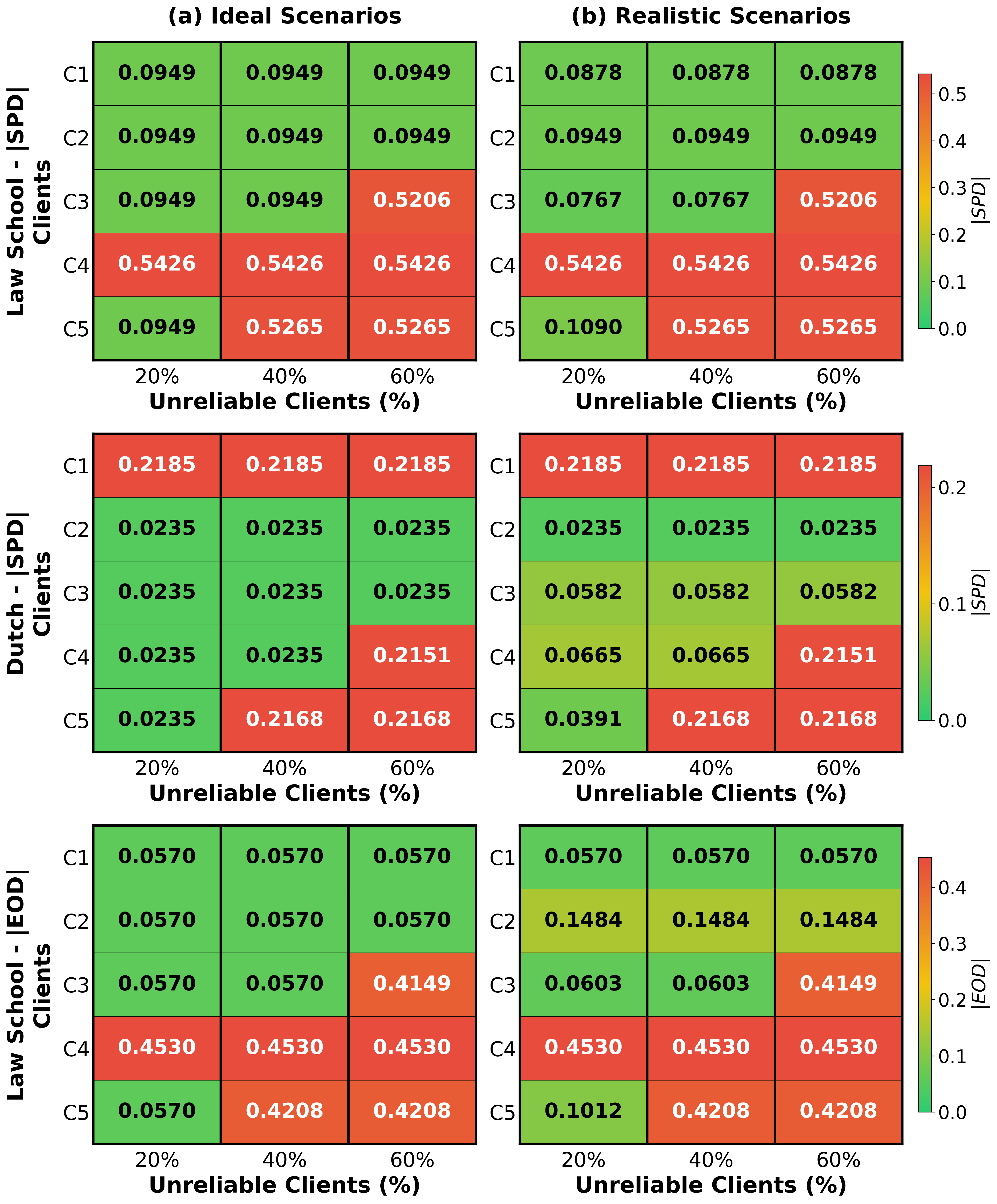}
  \caption{Client-level fairness measures of proxy data for different unreliable client percentages, different scenarios, and different fairness metrics.}
  \label{fig:clients_syn_fairness_heatmap}
\end{figure}

Figure~\ref{fig:clients_syn_fairness_heatmap} summarizes the client-level proxy data fairness measures across all experimental scenarios for a given random seed setting, for illustration purposes. Each value in the heatmap represents the fairness metric of the corresponding client's proxy data passed to the server. Cells with smaller values (more green) correspond to more reliable clients, while larger fairness values (more red) correspond to unreliable clients whose proxy data exhibit substantial bias. Each column within a subplot corresponds to a specific scenario configuration. Note that we do not evaluate the \texttt{Dutch} dataset under the EO measure because the raw dataset already achieves $|\text{EOD}| = 0.02$ leaving little room for improvement.

For the server-side root dataset, we allocate (at random) 0.5\% of each client's local original training data, which is used 
in the bilevel optimization framework.

\subsection{Baselines and Existing Robust Approaches}

Our baseline method (\textit{Baseline}) aggregates all client proxy data and trains a global model without considering fairness.


We compare with two existing robust approaches in FL, namely FedASL~\cite{talukder2022computationally} and FedNolowe~\cite{le2025fednolowe}. Both methods compute client-level aggregation weights based on client loss values but do not incorporate fairness, as they were designed for robustness against data heterogeneity in distributed learning. Importantly, these methods implicitly assume unreliable clients remain in the minority; when this assumption is violated, their weighting mechanisms may fail. These methods also rely on iterative communication to adaptively refine aggregation weights. To adapt them to our one-shot CML setting, we obtain fully-trained local models by training on each client's proxy data at the server, compute the corresponding client losses, and then apply their weighting schemes once to aggregate these models into a global model. This adaptation ensures a fair comparison under the one-shot communication constraint. FedASL defines a ``good region'' based on the median and standard deviation of client losses; clients whose losses fall within this region receive higher weights, and those outside receive lower weights. Following the original paper, we set $\alpha = 0.9$ (related to the good region definition) and $\beta = 0.2$ (related to weight preference within the good region). FedNolowe normalizes client losses and assigns higher weights to clients with smaller losses, with weights varying linearly across clients; no hyperparameter tuning is needed. All methods are evaluated under the same data partition and proxy data generation protocol to ensure a consistent comparison.

\subsection{Experimental Results and Discussion}

\begin{table}[t]
\centering
\caption{Comparison of final client weights on the \texttt{Law School} and \texttt{Dutch} datasets for the ideal scenarios in Section~\ref{subsec:scenarios}}
\label{tab:weights_comparison}
\begin{threeparttable}
\footnotesize
\begin{tabular}{llccccc}
\toprule
\textbf{Scenarios} & \textbf{Methods} & \multicolumn{5}{c}{\textbf{Weights}} \\
\midrule
& & C1 & C2 & C3 & \textbf{C4*} & C5 \\
\cmidrule(l){3-7}
\multirow{3}{*}{Law, SP, 20\%}
& Ours      & 0.25 & 0.25 & 0.25 & 0.00 & 0.25 \\
& FedASL    & 0.25 & 0.25 & 0.25 & 0.02 & 0.25 \\
& FedNolowe & 0.21 & 0.21 & 0.21 & 0.15 & 0.21 \\
\midrule
& & C1 & C2 & C3 & \textbf{C4*} & \textbf{C5*} \\
\cmidrule(l){3-7}
\multirow{3}{*}{Law, SP, 40\%}
& Ours      & 0.33 & 0.33 & 0.33 & 0.00 & 0.00 \\
& FedASL    & 0.31 & 0.31 & 0.31 & 0.04 & 0.03 \\
& FedNolowe & 0.22 & 0.22 & 0.22 & 0.17 & 0.16 \\
\midrule
& & C1 & C2 & \textbf{C3*} & \textbf{C4*} & \textbf{C5*} \\
\cmidrule(l){3-7}
\multirow{3}{*}{Law, SP, 60\%}
& Ours      & 0.50 & 0.50 & 0.00 & 0.00 & 0.00 \\
& FedASL    & 0.04 & 0.04 & 0.31 & 0.31 & 0.31 \\
& FedNolowe & 0.23 & 0.23 & 0.18 & 0.19 & 0.18 \\
\midrule[\heavyrulewidth]
& & \textbf{C1*} & C2 & C3 & C4 & C5 \\
\cmidrule(l){3-7}
\multirow{3}{*}{Dutch, SP, 20\%}
& Ours      & 0.00 & 0.25 & 0.25 & 0.25 & 0.25 \\
& FedASL    & 0.02 & 0.25 & 0.25 & 0.25 & 0.25 \\
& FedNolowe & 0.19 & 0.20 & 0.20 & 0.20 & 0.20 \\
\midrule
& & \textbf{C1*} & C2 & C3 & C4 & \textbf{C5*} \\
\cmidrule(l){3-7}
\multirow{3}{*}{Dutch, SP, 40\%}
& Ours      & 0.00 & 0.33 & 0.33 & 0.33 & 0.00 \\
& FedASL    & 0.03 & 0.31 & 0.31 & 0.31 & 0.03 \\
& FedNolowe & 0.19 & 0.20 & 0.20 & 0.20 & 0.19 \\
\midrule
& & \textbf{C1*} & C2 & C3 & \textbf{C4*} & \textbf{C5*} \\
\cmidrule(l){3-7}
\multirow{3}{*}{Dutch, SP, 60\%}
& Ours      & 0.00 & 0.50 & 0.50 & 0.00 & 0.00 \\
& FedASL    & 0.31 & 0.03 & 0.03 & 0.31 & 0.31 \\
& FedNolowe & 0.20 & 0.21 & 0.21 & 0.20 & 0.20 \\
\midrule[\heavyrulewidth]
& & C1 & C2 & C3 & \textbf{C4*} & C5 \\
\cmidrule(l){3-7}
\multirow{3}{*}{Law, EO, 20\%}
& Ours      & 0.25 & 0.25 & 0.25 & 0.00 & 0.25 \\
& FedASL    & 0.25 & 0.25 & 0.25 & 0.02 & 0.25 \\
& FedNolowe & 0.22 & 0.22 & 0.22 & 0.13 & 0.22 \\
\midrule
& & C1 & C2 & C3 & \textbf{C4*} & \textbf{C5*} \\
\cmidrule(l){3-7}
\multirow{3}{*}{Law, EO, 40\%}
& Ours      & 0.33 & 0.33 & 0.33 & 0.00 & 0.00 \\
& FedASL    & 0.31 & 0.31 & 0.31 & 0.04 & 0.03 \\
& FedNolowe & 0.23 & 0.23 & 0.23 & 0.17 & 0.15 \\
\midrule
& & C1 & C2 & \textbf{C3*} & \textbf{C4*} & \textbf{C5*} \\
\cmidrule(l){3-7}
\multirow{3}{*}{Law, EO, 60\%}
& Ours      & 0.50 & 0.50 & 0.00 & 0.00 & 0.00 \\
& FedASL    & 0.04 & 0.04 & 0.31 & 0.31 & 0.31 \\
& FedNolowe & 0.23 & 0.23 & 0.18 & 0.19 & 0.17 \\
\bottomrule
\end{tabular}
\begin{tablenotes}
\footnotesize
\item \textit{Note:} Each number represents the final aggregation weight assigned to each client (C1--C5). Client labels with bold asterisk (*) indicate unreliable clients. The column ``Scenarios'' gives the dataset, fairness metric (SP or EO), and unreliable client percentage~(\%).\!\!
\end{tablenotes}
\end{threeparttable}
\end{table}

\begin{table}[t]
\centering
\caption{Performance comparison between our method and the baseline method for the Realistic Scenarios described in Section~\ref{subsec:scenarios}.}
\label{tab:results}
\begin{threeparttable}
\footnotesize
\setlength{\tabcolsep}{2pt}
\begin{tabular}{@{}cccccccc@{}}
\toprule
\multirow{2}[2]{*}{\makecell{\textbf{Dataset}}} &
\multirow{2}[2]{*}{\makecell{\textbf{Fairness}\\\textbf{Metric}}} &
\multirow{2}[2]{*}{\makecell{\textbf{Unreliable}\\\textbf{Clients(\%)}}} &
\multirow{2}[2]{*}{$\rho_s$} &
\multicolumn{2}{c}{\textbf{Baseline}} &
\multicolumn{2}{c}{\textbf{Ours}} \\
\cmidrule(lr){5-6} \cmidrule(lr){7-8}
& & & & \textbf{Acc(\%)} & \textbf{Fair} & \textbf{Acc(\%)} & \textbf{Fair} \\
\midrule
\multirow{6}[2]{*}{\makecell{Law\\School}} & \multirow{6}[2]{*}{$|$SPD$|$}
& \multirow{2}{*}{20\%}
  & $0$       & 88.21 & 0.1465 & 88.25 & 0.1612 \\
& & & Adaptive  &       &        & 84.52 & 0.0369 \\
\cmidrule(lr){3-8}
& & \multirow{2}{*}{40\%}
  & $0$       & 88.49 & 0.2254 & 88.50 & 0.2483 \\
& & & Adaptive  &       &        & 85.07 & 0.0382 \\
\cmidrule(lr){3-8}
& & \multirow{2}{*}{60\%}
  & $0$       & 88.64 & 0.3133 & 88.57 & 0.3132 \\
& & & Adaptive  &       &        & 84.72 & 0.0403 \\
\midrule
\multirow{6}[2]{*}{Dutch} & \multirow{6}[2]{*}{$|$SPD$|$}
& \multirow{2}{*}{20\%}
  & $0$       & 77.18 & 0.0802 & 77.23 & 0.0833 \\
& & & Adaptive  &       &        & 76.91 & 0.0519 \\
\cmidrule(lr){3-8}
& & \multirow{2}{*}{40\%}
  & $0$       & 77.55 & 0.1118 & 77.59 & 0.1127 \\
& & & Adaptive  &       &        & 76.99 & 0.0530 \\
\cmidrule(lr){3-8}
& & \multirow{2}{*}{60\%}
  & $0$       & 77.84 & 0.1393 & 77.82 & 0.1346 \\
& & & Adaptive  &       &        & 76.76 & 0.0442 \\
\midrule
\multirow{6}[2]{*}{\makecell{Law\\School}} & \multirow{6}[2]{*}{$|$EOD$|$}
& \multirow{2}{*}{20\%}
  & $0$       & 88.63 & 0.1130 & 88.66 & 0.1145 \\
& & & Adaptive  &       &        & 81.32 & 0.0120 \\
\cmidrule(lr){3-8}
& & \multirow{2}{*}{40\%}
  & $0$       & 88.65 & 0.1597 & 88.63 & 0.1676 \\
& & & Adaptive  &       &        & 81.84 & 0.0174 \\
\cmidrule(lr){3-8}
& & \multirow{2}{*}{60\%}
  & $0$       & 88.57 & 0.2211 & 87.85 & 0.3154 \\
& & & Adaptive  &       &        & 81.54 & 0.0255 \\
\bottomrule
\end{tabular}
\begin{tablenotes}
\footnotesize
\item \textit{Note:} The column ``Fair'' denotes the absolute value of the metric specified in the Fairness Metric column; the column Unreliable Clients gives the percentage of unreliable clients. ``Adaptive'' denotes a scheduled $\rho_s$ as described in Section~\ref{sec:implm.details}.
\end{tablenotes}
\end{threeparttable}
\end{table}

\begin{table}[t]
\centering
\caption{Performance comparison for the  realistic scenarios}
\label{tab:performance}
\begin{threeparttable}
\scriptsize
\setlength{\tabcolsep}{1.5pt}
\begin{tabular}{@{}ccccccccc@{}}
\toprule
\multirow{2}[2]{*}{\makecell{\textbf{Dataset}}} & \multirow{2}[2]{*}{\makecell{\textbf{Fairness}\\\textbf{Metric}}} & \multirow{2}[2]{*}{\makecell{\textbf{Unreliable}\\\textbf{Clients(\%)}}} & \multicolumn{2}{c}{\textbf{Ours}} & \multicolumn{2}{c}{\textbf{FedASL}} & \multicolumn{2}{c}{\textbf{FedNolowe}} \\
\cmidrule(lr){4-5} \cmidrule(lr){6-7} \cmidrule(lr){8-9}
& & & \textbf{Acc(\%)} & \textbf{Fair} & \textbf{Acc(\%)} & \textbf{Fair} & \textbf{Acc(\%)} & \textbf{Fair} \\
\midrule
\multirow{3}{*}{\makecell{Law\\School}} & \multirow{3}{*}{$|$SPD$|$}
& 20\% & 84.52 & 0.0369 & 85.20 & 0.0397 & 86.57 & 0.0867 \\
& & 40\% & 85.07 & 0.0382 & 85.77 & 0.0567 & 87.89 & 0.1534 \\
& & 60\% & 84.72 & 0.0403 & 87.94 & 0.4948 & 88.23 & 0.2574 \\
\midrule
\multirow{3}{*}{Dutch} & \multirow{3}{*}{$|$SPD$|$}
& 20\% & 76.91 & 0.0519 & 77.12 & 0.0606 & 77.33 & 0.0786 \\
& & 40\% & 76.99 & 0.0530 & 77.41 & 0.0899 & 77.78 & 0.1108 \\
& & 60\% & 76.76 & 0.0442 & 78.57 & 0.2092 & 78.09 & 0.1378 \\
\midrule
\multirow{3}{*}{\makecell{Law\\School}} & \multirow{3}{*}{$|$EOD$|$}
& 20\% & 81.32 & 0.0120 & 82.48 & 0.0150 & 85.24 & 0.0427 \\
& & 40\% & 81.84 & 0.0174 & 83.92 & 0.0188 & 87.61 & 0.1048 \\
& & 60\% & 81.54 & 0.0255 & 87.86 & 0.3735 & 88.22 & 0.1768 \\
\bottomrule
\end{tabular}
\begin{tablenotes}
\footnotesize
\item \textit{Note:} The column ``Fair'' denotes the absolute value of the metric specified in the Fairness Metric column. Results for ``Ours" use the adaptive $\rho_s$ schedule.
\end{tablenotes}
\end{threeparttable}
\end{table}

\noindent\textbf{Analysis of Client Weight Assignment.}
We first analyze the learned client weights under the ideal scenarios described in Section~\ref{subsec:scenarios} to examine the behavior of the proposed weighting mechanism. Using the \texttt{Law School} and \texttt{Dutch} datasets as representative cases, Table~\ref{tab:weights_comparison} summarizes the final client weights for both settings.
In this table, consider the scenario when SP is used and there are $20\%$ unreliable clients, as an illustrative example. For our method, 
the final weights are $[0.25, 0.25, 0.25, 0, 0.25]$, assigning equal weights to reliable clients and zero weight to the unreliable one. FedASL behaves similarly when unreliable clients are the minority, but fails once they exceed 50\%, as the median-based good region becomes dominated by unreliable clients (e.g., $[0.04, 0.04, 0.31, 0.31, 0.31]$ for the 60\% case). FedNolowe assigns weights linearly based on normalized client losses, limiting its ability to downweight unreliable clients (e.g., $[0.23, 0.23, 0.18, 0.19, 0.18]$ for the 60\% case).

\smallskip
\noindent
\textbf{Comparison with the baseline.} Table~\ref{tab:results} summarizes the performance between our method and the baseline method for different values of the penalty parameter $\rho_s$. When $\rho_s = 0$, no fairness constraint is enforced, and our method exhibits high fairness metric values, indicating unfairness in the global model. This is expected, as the optimization focuses solely on minimizing loss without any fairness consideration. As $\rho_s$ adaptively increases, the fairness constraints become more stringent. For example, on \texttt{Dutch} under the SP metric with 60\% unreliable clients and the realistic scenarios, our method achieves $|\text{SPD}|=0.0442$, compared to 0.1393 for the baseline, with accuracy dropping only from 77.84\% to 76.76\%. Compared to the baseline, our method reduces unfairness by more than 68\%. 
Similar trends are observed for other datasets, fairness metrics, and unreliable client percentages.

\begin{figure}[t]
  \centering
  \includegraphics[width=\columnwidth]{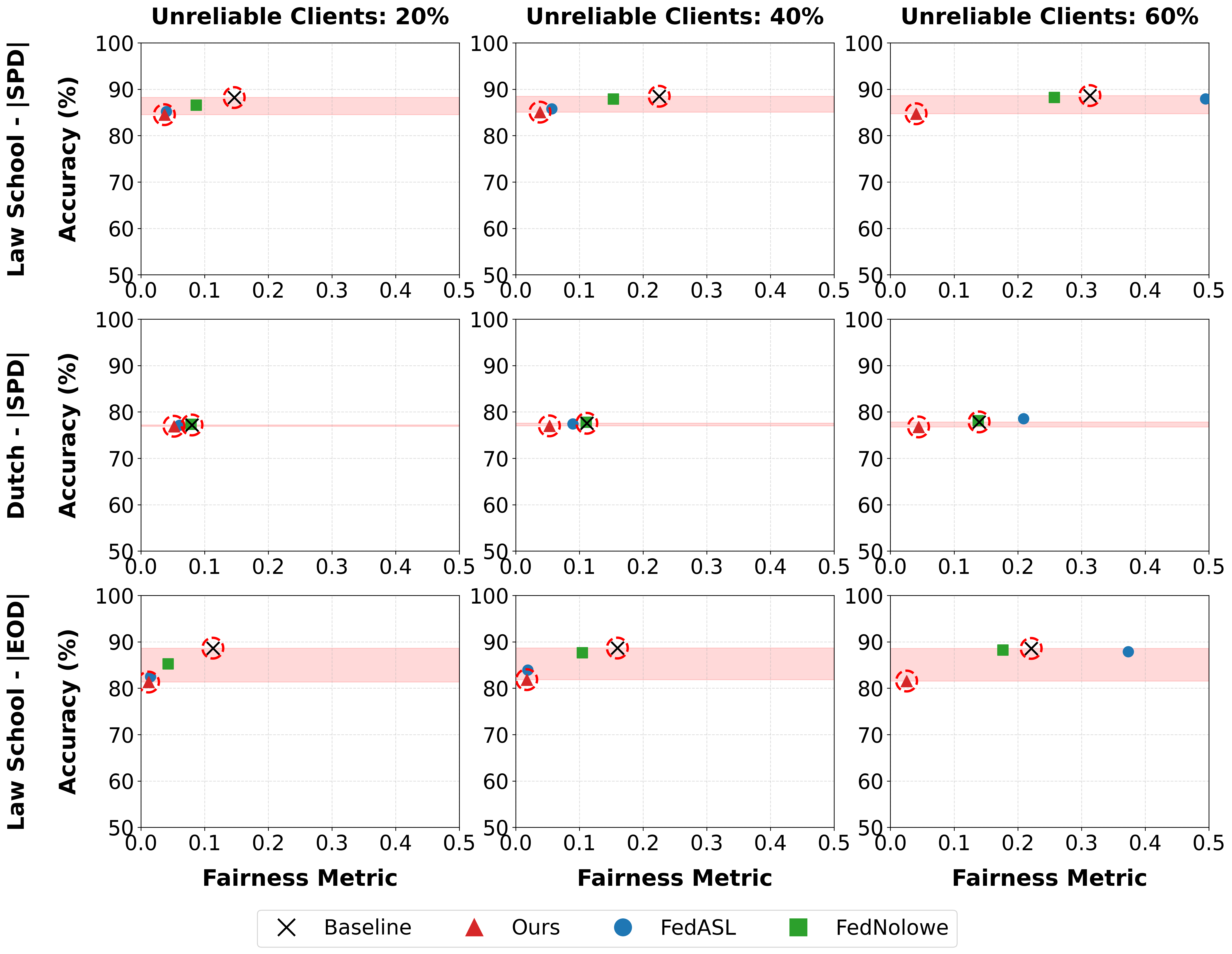}
  \caption{Bi-criteria comparison of accuracy and fairness across different methods for the realistic scenario setting discussed in Section~\ref{subsec:scenarios}.}
  \label{fig:bi_criteria}
\end{figure}

\smallskip
\noindent
\textbf{Comparison with Existing Robust Methods.} Table~\ref{tab:performance} compares the performance of FedASL and FedNolowe. 
Figure~\ref{fig:bi_criteria} visualizes this trade-off for the realistic scenarios (see Section~\ref{subsec:scenarios}), where each point represents the average over five runs with different random seed settings for a method's accuracy and fairness metric. Points closer to the top-left corner have a better accuracy-fairness trade-off. We highlight our method and the baseline with red dashed circles to facilitate comparison. The red shaded band is the accuracy range that our method operates while improving fairness; a narrower band indicates less accuracy is sacrificed. For the 20\% and 40\% unreliable client settings, our method achieves comparable fairness to FedASL, with both methods outperforming FedNolowe by a substantial margin. However, when unreliable clients dominate the system (60\%), FedASL suffers significant fairness degradation, whereas our method remains effective. For instance, on \texttt{Law School} under the SP metric with 60\% unreliable clients, our method achieves $|\text{SPD}| = 0.0403$ compared to 0.4948 for FedASL and 0.2574 for FedNolowe.


\section{Conclusion}


This paper addresses fairness degradation caused by unreliable clients in one-shot CML, where the single-round communication limits server-side correction. We proposed a bilevel optimization framework that learns client-level aggregation weights while enforcing fairness constraints using a small trusted root dataset as a server-side defense. Unlike robustness-only methods, our fairness-aware reweighting remains stable even when unreliable clients dominate the system, substantially improving group fairness with limited accuracy loss and consistently outperforming existing weighting-based aggregation methods. This reliance on a root dataset is a limitation that may not be feasible in all scenarios.



\bibliographystyle{IEEEtran}
\bibliography{refs}


\end{document}